\documentclass[acmtog]{acmart}
\PassOptionsToPackage{table}{xcolor}

\usepackage{booktabs} 

\usepackage[ruled]{algorithm2e} 

\SetAlFnt{\small}
\SetAlCapFnt{\small}
\SetAlCapNameFnt{\small}
\SetAlCapHSkip{0pt}

\copyrightyear{2026}
\acmYear{2026}
\setcopyright{cc}
\setcctype{by-nc-nd}
\acmConference[SIGGRAPH Conference Papers '26]{Special Interest Group on Computer Graphics and Interactive Techniques Conference Conference Papers}{July 19--23, 2026}{Los Angeles, CA, USA}
\acmBooktitle{Special Interest Group on Computer Graphics and Interactive Techniques Conference Conference Papers (SIGGRAPH Conference Papers '26), July 19--23, 2026, Los Angeles, CA, USA}
\acmDOI{10.1145/3799902.3811163}
\acmISBN{979-8-4007-2554-8/2026/07}

\usepackage{colortbl}
\usepackage{graphicx}
\usepackage{rotating}

\begin{document}
\title{Ref-DGS: Reflective Dual Gaussian Splatting}


\author{Ningjing Fan}
\orcid{0009-0003-5546-892X}
\affiliation{\institution{Chongqing University}
\city{Chongqing}
\country{China}}
\email{njfancalm@gmail.com}

\author{Yiqun Wang}
\authornote{Corresponding author.}
\orcid{0000-0003-1942-5597}
\affiliation{\institution{Chongqing University}
\city{Chongqing}
\country{China}}
\email{csyqwang@hotmail.com}

\author{Dong-Ming Yan}
\orcid{0000-0003-2209-2404}
\affiliation{\institution{MAIS, Institute of Automation, Chinese Academy of Sciences and UCAS}
\city{Beijing}
\country{China}}
\email{yandongming@gmail.com}

\author{Peter Wonka}
\orcid{0000-0003-0627-9746}
\affiliation{\institution{King Abdullah University of Science and Technology (KAUST)}
\city{Thuwal}
\country{Saudi Arabia}}
\email{pwonka@gmail.com}

\renewcommand\shortauthors{Fan et al.}

\begin{abstract}
The reflective appearance, especially strong and typically near-field specular reflections, poses a fundamental challenge for accurate surface reconstruction and novel view synthesis. Existing Gaussian splatting methods either fail to model near-field specular reflections or rely on explicit ray tracing at substantial computational cost. We present \textbf{Ref-DGS}, a reflective dual Gaussian splatting framework that addresses this trade-off by decoupling surface reconstruction from specular reflection within an efficient rasterization-based pipeline. Ref-DGS introduces a dual Gaussian scene representation consisting of geometry Gaussians and complementary local reflection Gaussians that capture near-field specular interactions without explicit ray tracing, along with a global environment reflection field for modeling far-field specular reflections. To predict specular radiance, we further propose a lightweight, physically-aware specular adaptive mixing shader that fuses global and local specular features. Experiments demonstrate that Ref-DGS achieves state-of-the-art performance on reflective scenes while training substantially faster than ray-based Gaussian methods.
\end{abstract}

%
%


%
%



\begin{teaserfigure}
  \centering
  \includegraphics[width=\textwidth]{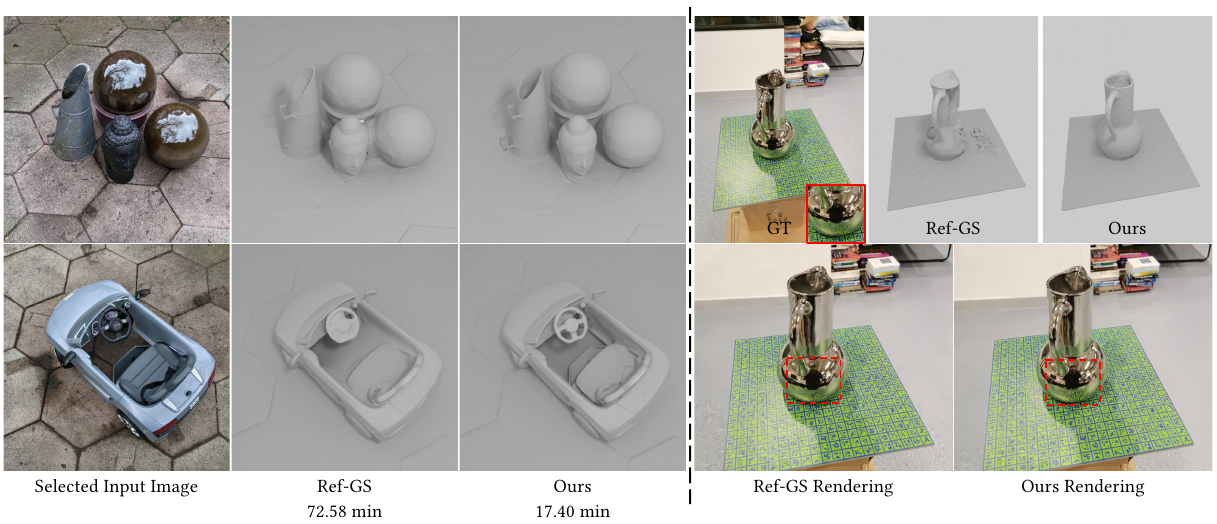}
  \vspace{-0.7cm}
  \caption{
    \textbf{Left:} Surface reconstruction results on the RefReal dataset for two challenging scenes, gardenspheres (top) and toycar (bottom), both with strong near-field specular reflections. The bottom row reports the average training time. 
    \textbf{Right:} Results on the GlossyReal dataset for the vase scene, showing surface reconstruction and rendering quality. Our method produces more accurate geometry and higher-quality reflections, as highlighted in the close-up regions.
    }
  \label{fig:teaser}
\end{teaserfigure}

\maketitle

\section{Introduction}
Scenes with reflective materials pose significant challenges for surface reconstruction and novel view synthesis, as view-dependent specular and indirect effects violate the multi-view appearance consistency assumptions commonly used for geometry recovery and rendering. While recent neural rendering methods achieve impressive results for predominantly diffuse scenes, faithfully handling specular reflections and view-dependent effects under complex lighting remains difficult, especially when high visual quality and practical efficiency are both required. NeRF-based approaches address view-dependent appearance through volumetric radiance fields and physically inspired shading~\cite{mildenhall2020nerf}, but their reliance on dense ray sampling and volumetric integration leads to high computational cost and limited scalability. In contrast, 3D Gaussian Splatting (3DGS) has emerged as an efficient alternative, enabling fast novel view synthesis via differentiable rasterization~\cite{kerbl20233d}. However, extending Gaussian-based representations to jointly support accurate surface reconstruction and faithful modeling of reflective effects within an efficient rasterization-based framework remains an open problem.

At the core of this difficulty lies a mismatch between the physical nature of specular reflection and the representations used in existing Gaussian-based methods. A large body of work~\cite{Huang2DGS2024, chen2024pgsr, Yu2024GOF, guedon2025milo} focuses on improving surface reconstruction quality, achieving accurate and stable geometry but without explicitly modeling specular reflection. In reflective scenes, view-dependent specular effects are therefore absorbed into the geometry during optimization, leading to surface shrinkage, local dents, and the loss of high-frequency details. Other approaches introduce explicit specular modeling within a Gaussian splatting framework~\cite{Jiang_2024_CVPR, yang2024spec, ye20243d, zhang2025ref}, typically by querying environment illumination or using single-bounce reflection models. While effective for far-field reflections, these models fail to capture near-field specular reflections, such as self-reflections and geometry-induced inter-reflections, which are often prominent on glossy surfaces. More recent methods~\cite{gao2024relightable, ICLR2025_abf3682c, Xie_2025_CVPR, zhangmaterialrefgs, zhu_2025_gsror} address this limitation by incorporating ray tracing to model near-field inter-reflections. Although physically accurate, this strategy incurs substantial computational overhead and significantly slows training, eroding the efficiency advantages that make Gaussian splatting attractive. Fundamentally, these limitations stem from attempting to encode both view-invariant geometry and complex view-dependent specular reflection within a single representation, forcing incompatible phenomena to be explained by the same parameters.

In this paper, we present \textbf{Ref-DGS}, a reflective dual Gaussian splatting framework that addresses two fundamental limitations of existing Gaussian-based rendering methods. First, we show that reflective appearance should not be treated as a geometric property: forcing view-dependent specular effects into surface-aligned Gaussians leads to unstable geometry and degraded surface reconstruction. Second, we demonstrate that explicitly separating specular reflection from geometry does not require costly ray tracing to model near-field inter-reflections within objects, which significantly undermines the efficiency advantages of Gaussian splatting. To resolve these issues, Ref-DGS introduces a dual Gaussian scene representation that assigns geometry and near-field specular reflections to two complementary sets of Gaussians, enabling reflective appearance modeling within an efficient rasterization-based pipeline. Far-field specular reflections are modeled via environment-based illumination and encoded as global specular features, while near-field specular reflections are captured using rasterization-friendly local reflection Gaussians and encoded as local specular features, which are further integrated through a lightweight, physically-aware specular adaptive mixing shader. This design preserves the efficiency and stability of Gaussian splatting while enabling faithful reflective appearance modeling, achieving state-of-the-art performance in both surface reconstruction and novel view synthesis, with substantially faster training than ray-based Gaussian methods.

In summary, our contributions are:
\begin{itemize}
    \item A reflective dual Gaussian splatting framework that introduces a \emph{dual Gaussian scene representation} to explicitly disentangle near-field specular reflections from surface geometry, assigning local reflective interactions to a dedicated set of local reflection Gaussians and enabling efficient reflective appearance modeling without explicit ray tracing.
    \item A \emph{global--local specular reflection representation} that models specular appearance using two complementary components: global specular features capturing far-field environment illumination, and local specular features extracted from the local reflection Gaussians to represent near-field inter-reflections.
    \item A lightweight \emph{physically-aware specular adaptive mixing shader} that fuses global and local specular features conditioned on material roughness and geometric angular factors, enabling fast convergence and state-of-the-art performance in both surface reconstruction and novel view synthesis.
\end{itemize}

\section{Related Work}
We review prior work on novel view synthesis and surface reconstruction, with a focus on neural and Gaussian-based scene representations and their treatment of reflective appearance.

\subsection{Novel View Synthesis}

Novel view synthesis aims to render photorealistic images from unseen viewpoints. NeRF-based approaches~\cite{mildenhall2020nerf, Barron_2021_ICCV, Barron_2022_CVPR} achieve high-quality view synthesis but suffer from costly volumetric ray marching. To improve efficiency, factorized or explicit scene representations have been proposed, including Plenoxels~\cite{Fridovich-Keil_2022_CVPR}, TensoRF~\cite{chen2022tensorf}, and Instant-NGP~\cite{mueller2022instant}. Recent efforts have focused on reflective view synthesis. NeRF-Casting~\cite{verbin2024nerf} improves specular rendering by tracing reflection rays through a NeRF representation. NeRFReN~\cite{guo2022nerfren} models reflective scenes using separate transmitted and reflected radiance fields, especially for mirror- and glass-like reflections.

3D Gaussian Splatting (3DGS)~\cite{kerbl20233d} further improves efficiency by representing scenes with explicit Gaussian primitives and enabling fast rasterization-based rendering. Building upon this framework, subsequent works extend Gaussian splatting to better handle reflective appearance and complex illumination in novel view synthesis, by incorporating view-dependent components, near-field illumination modeling, deferred shading, or explicit reflective transport~\cite{Jiang_2024_CVPR, yang2024spec, Liang_2024_CVPR, ye20243d, zhang2025ref, gao2024relightable, ICLR2025_abf3682c, Xie_2025_CVPR, zhangmaterialrefgs}. These approaches improve visual fidelity under challenging lighting conditions, at the cost of increased computational complexity.
In addition, many surface reconstruction works reviewed in the following subsection also support reflective novel view synthesis.

\subsection{Surface Reconstruction}

\subsubsection{Volumetric surface reconstruction with neural fields.}
Surface reconstruction is a core problem in neural rendering. NeRF-based methods and implicit signed distance field (SDF) formulations represent scenes as continuous volumetric fields and have been widely adopted for accurate geometry reconstruction~\cite{mildenhall2020nerf, wang2021neus}. Several extensions incorporate reflective appearance and view-dependent effects through physically inspired shading or radiance decomposition~\cite{verbin2022refnerf, liang2023envidr, ge2023ref, liu2023nero, li2024tensosdf, wang2024unisdf, fan2025factored}. Despite their high reconstruction quality, these methods rely on dense ray sampling and volumetric integration, resulting in high computational cost and limited scalability.

\subsubsection{Gaussian-based surface reconstruction in predominantly diffuse scenes.}
Gaussian-based representations provide an efficient alternative to volumetric formulations for surface reconstruction. SuGaR~\cite{Guedon_2024_CVPR}, PGSR~\cite{chen2024pgsr}, and GOF~\cite{Yu2024GOF} improve geometric fidelity within the Gaussian framework through enhanced depth estimation and geometric regularization. Other approaches, such as NeuSG~\cite{chen2023neusg}, GSDF~\cite{NEURIPS2024_ea13534e}, and 3DGSR~\cite{lyu20243dgsr}, integrate Gaussian representations with signed distance fields to enable explicit surface reconstruction. 
In parallel, surface-aware Gaussian formulations progressively enhance geometric expressiveness, where 2D Gaussian Splatting (2DGS)~\cite{Huang2DGS2024} introduces local surface parameterization, and Quadratic Gaussian Splatting (QGS)~\cite{Zhang_2025_ICCV} further extends this formulation with deformable quadric primitives and geometry-aware density modeling to better capture complex surface curvature. In a different direction, MILo~\cite{guedon2025milo} jointly optimizes Gaussians and an explicitly extracted mesh by enforcing bidirectional consistency between the two representations, enabling synchronized geometry refinement without introducing an additional SDF representation.

\subsubsection{Gaussian-based surface reconstruction in reflective scenes.}
Surface reconstruction becomes more challenging in the presence of reflective appearance, as view-dependent specular and indirect effects violate appearance consistency assumptions commonly used for geometry recovery. To address this, several works extend Gaussian splatting to better account for reflective effects during reconstruction and rendering. GaussianShader~\cite{Jiang_2024_CVPR} and Spec-Gaussian~\cite{yang2024spec} introduce explicit specular components within the Gaussian framework, primarily targeting far-field or direct reflections. Deferred rendering formulations further decouple geometry and shading to improve reconstruction robustness under complex illumination, as explored in 3DGS-DR~\cite{ye20243d} and Ref-GS~\cite{zhang2025ref}. 
To capture near-field inter-reflections, GS-IR~\cite{Liang_2024_CVPR} represents reflective transport using spherical harmonic features, while Ref-Gaussian~\cite{ICLR2025_abf3682c}, EnvGS~\cite{Xie_2025_CVPR}, and MaterialRefGS~\cite{zhangmaterialrefgs} incorporate explicit ray tracing to model near-field reflective interactions, at the cost of increased computational overhead. GS-ROR$^2$ further addresses surface reconstruction in reflective scenes by bidirectionally coupling 3D Gaussian splatting with signed distance fields, leveraging mutual depth and normal supervision to jointly enable efficient relighting and high-quality geometry reconstruction. Overall, these approaches explore different trade-offs between geometric fidelity, reflective expressiveness, and computational efficiency.

\section{Method}
\begin{figure*}
\centering
\includegraphics[width=\linewidth]{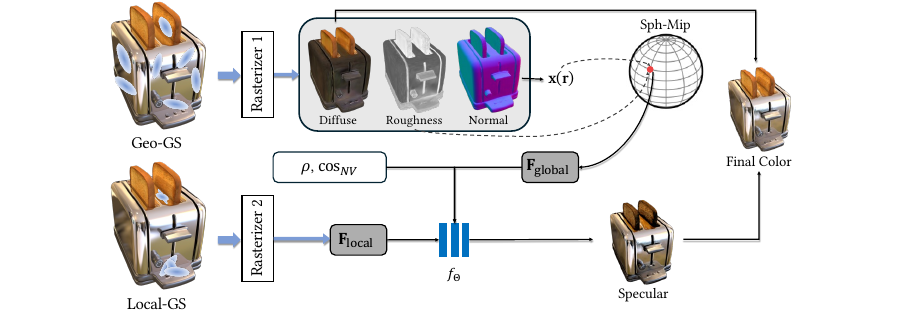}
\vspace{-0.9cm}
\caption{An overview of our Ref-DGS framework.}
\vspace{-0.3cm}
\label{fig:pipeline}
\end{figure*}

This section presents the proposed reflective dual Gaussian splatting framework. 
We first review 2D Gaussian splatting as the foundation of our approach (Sec.~\ref{sec:2dgs}). We then describe the dual Gaussian splatting framework, including the dual Gaussian scene representation (Sec.~\ref{sec:dual_gaussian}), the construction of global and local specular features (Sec.~\ref{sec:global_local}), and a physically-aware specular adaptive mixing shader (Sec.~\ref{sec:shader}).

\subsection{Gaussian Splatting}
\label{sec:2dgs}

Our method builds upon 2D Gaussian Splatting (2DGS)~\cite{huang20242d}, which represents a scene as a collection of planar Gaussian surface primitives and renders them using a fast, fully differentiable splatting rasterizer. Compared to vanilla 3DGS~\cite{kerbl20233d}, 2DGS constrains each primitive to a local tangent disk, yielding a surface-aligned representation that is better suited for thin structures and for producing reliable geometric information such as depth and surface normals.

Each surface primitive is parameterized by a 3D center $\mathbf{p}_k \in \mathbb{R}^3$, an opacity $\alpha_k \in (0,1)$, and a local planar Gaussian defined by a rotation matrix $\mathbf{R}_k = [\mathbf{t}_{u,k}, \mathbf{t}_{v,k}, \mathbf{t}_{w,k}]$, where $\mathbf{t}_{u,k}$ and $\mathbf{t}_{v,k}$ are orthogonal tangential directions and $\mathbf{t}_{w,k} = \mathbf{t}_{u,k} \times \mathbf{t}_{v,k}$, and a diagonal scaling matrix $\mathbf{S}_k = \mathrm{diag}(s_{u,k}, s_{v,k}, 0)$. A point on the local tangent disk is expressed in local coordinates as
\begin{equation}
\mathbf{P}_k(u,v)=\mathbf{p}_k + s_{u,k}\mathbf{t}_{u,k}u + s_{v,k}\mathbf{t}_{v,k}v,
\end{equation}
where $(u,v)$ denote normalized local coordinates on the disk. The corresponding Gaussian weight is defined as a unit-variance isotropic Gaussian in the local parameter space,
\begin{equation}
\mathcal{G}(u,v)=\exp\!\left(-\frac{u^2+v^2}{2}\right).
\end{equation}

Following the standard splatting formulation~\cite{kerbl20233d, huang20242d}, each planar Gaussian projects to an anisotropic ellipse in screen space through perspective projection and local linearization. The ellipse is evaluated within a conservative bounding box and organized into screen-space tiles for efficient rasterization. Primitives are depth-sorted by their centers and composited in a front-to-back order using alpha compositing:
\begin{equation}
\mathbf{c}(\mathbf{q})=\sum_{k=1}^{K}\mathbf{c}_k\,\alpha_k\,\hat{\mathcal{G}}_k(\mathbf{q})
\prod_{j=1}^{k-1}\left(1-\alpha_j\,\hat{\mathcal{G}}_j(\mathbf{q})\right),
\end{equation}
where $\hat{\mathcal{G}}_k(\mathbf{q})$ denotes the projected 2D Gaussian kernel evaluated at pixel $\mathbf{q}$, and $\mathbf{c}_k$ represents a per-primitive attribute such as color or a learnable feature. Rendering terminates early once the accumulated opacity saturates, and the entire process is fully differentiable with respect to both geometric parameters and per-primitive attributes.

\subsection{Dual Gaussian Splatting}

We propose a dual Gaussian splatting framework that combines far-field specular reflections obtained by querying a learnable spherical feature map using the spherical mipmap (Sph-Mip) encoding with near-field specular reflections captured by an auxiliary set of Gaussians. The global and local specular features are subsequently fused by a physically-aware shader to predict specular radiance. 
An overview of our framework is shown in Fig.~\ref{fig:pipeline}.

\subsubsection{Dual Gaussian Scene Representation}
\label{sec:dual_gaussian}
Building on the 2D Gaussian splatting (Sec.~\ref{sec:2dgs}), we introduce a dual Gaussian scene representation that decomposes the scene into two complementary sets of Gaussians with distinct roles: the geometry Gaussians $\mathcal{G}_{\mathrm{geo}}$ and the local reflection Gaussians $\mathcal{G}_{\mathrm{local}}$.

\paragraph{Geometry Gaussians $\mathcal{G}_{\mathrm{geo}}$.}
The geometry Gaussians are designed to capture view-independent scene structure and to provide stable geometric information for rendering and surface reconstruction. 
Each primitive in $\mathcal{G}_{\mathrm{geo}}$ follows the 2DGS formulation and is augmented with material-related attributes, including diffuse color and surface roughness, which are treated as view-invariant properties tied to the underlying geometry. 
When splatted, $\mathcal{G}_{\mathrm{geo}}$ produces screen-space attribute maps such as diffuse color $\mathbf{C}_{\mathrm{diff}}(\mathbf{p})$ and roughness $\rho(\mathbf{p})$, together with geometric information including depth and surface normals. 
These maps are used as inputs to the subsequent shading stages and reflection modeling.

\paragraph{Local reflection Gaussians $\mathcal{G}_{\mathrm{local}}$.}
The local reflection Gaussians are introduced to model local, view-dependent specular reflection effects arising from nearby geometry. 
Each primitive in $\mathcal{G}_{\mathrm{local}}$ stores a learnable local specular feature $\mathbf{f}\!\in\!\mathbb{R}^{d}$. 
Splatting $\mathcal{G}_{\mathrm{local}}$ produces the corresponding screen-space feature map $\mathbf{F}_{\mathrm{local}}(\mathbf{p})$. 
Intuitively, $\mathbf{F}_{\mathrm{local}}$ provides a compact implicit representation of self-reflections and inter-reflections, enabling efficient modeling of near-field specular reflections within a rasterization-based and fully differentiable framework. It is important that these Gaussians do not influence and thereby distort the surface reconstruction, which is handled solely by the geometry Gaussians.

\subsubsection{Global--Local Specular Features}
\label{sec:global_local}
For each pixel $\mathbf{p}$, we construct specular features by combining (i) far-field global specular features obtained by querying a learnable spherical feature map using the spherical mipmap (Sph-Mip) encoding, and (ii) near-field local specular features rendered from the local reflection Gaussians $\mathcal{G}_{\mathrm{local}}$.

\paragraph{Far-field global specular reflection via Sph-Mip.}
Following Ref-GS~\cite{zhang2025ref}, we represent far-field specular reflections using a learnable spherical feature map $\mathbf{M}$. 
Global specular features are obtained by applying the Sph-Mip encoding (See the supplementary material Sec.~B for more details), which queries $\mathbf{M}$ conditioned on the reflection direction and surface roughness.
For each pixel, we compute the specular reflection direction $\mathbf{r}$ from the estimated surface normal and view direction, convert it to spherical coordinates $\mathbf{x}(\mathbf{r})\in[0,1]^2$, and select a roughness-dependent mipmap level:
\begin{equation}
\mathbf{F}_{\mathrm{global}}(\mathbf{p})
= \mathrm{Sph\text{-}Mip}\!\left(\mathbf{x}(\mathbf{r}),\; \ell\!\left(\rho(\mathbf{p})\right),\; \mathbf{M} \right),
\label{eq:global_feat}
\end{equation}
where $\mathbf{F}_{\mathrm{global}}(\mathbf{p}) \in \mathbb{R}^{d}$, and $\ell(\cdot)$ maps surface roughness to a continuous mipmap level to approximate prefiltered far-field specular reflections from the environment.

\paragraph{Near-field local specular reflection from $\mathcal{G}_{\mathrm{local}}$.}
We render the local reflection Gaussians $\mathcal{G}_{\mathrm{local}}$ to obtain per-pixel local specular features
\begin{equation}
\mathbf{F}_{\mathrm{local}}(\mathbf{p}) \in \mathbb{R}^{d},
\label{eq:local_feat}
\end{equation}
which captures near-field specular reflections that are difficult to represent using a global environment model alone.

The global and local specular features are subsequently fused by the specular shader described in Sec.~\ref{sec:shader}.

\begin{figure}
\centering
\includegraphics[width = \linewidth]{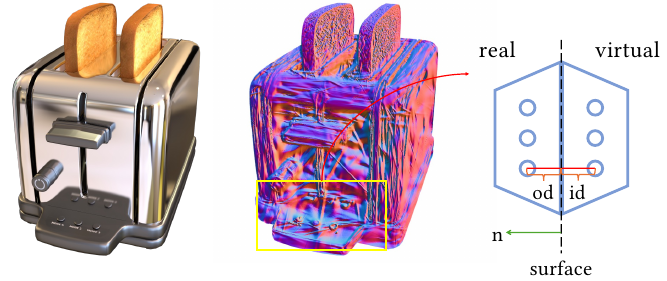}
\vspace{-0.7cm}
\caption{
\textbf{Left:} Input image.
\textbf{Middle:} Normal map of the local reflection Gaussians $\mathcal{G}_{\mathrm{local}}$.
\textbf{Right:} Illustration of specular reflection, where the surface normal $\mathbf{n}$ is perpendicular to the surface, and the image distance ($\mathrm{id}$) equals the object distance ($\mathrm{od}$).
}
\vspace{-0.3cm}
\label{fig:ref_normal}
\end{figure}

\paragraph{Discussion.}
We justify the necessity of the proposed local reflection Gaussians by analyzing the physical properties of specular reflections.
Attempting to encode local specular features directly into the geometry Gaussians $\mathcal{G}_{\mathrm{geo}}$ is problematic due to the depth discrepancy between the physical surface and the mirror-induced virtual image on a glossy surface. As illustrated in Fig.~\ref{fig:ref_normal} (right), according to the law of specular reflection for a planar mirror, the image distance ($\mathrm{id}$) equals the object distance ($\mathrm{od}$), implying that the virtual image lies "behind" the reflective surface. This effect is visualized in Fig.~\ref{fig:ref_normal} (middle), where the normal map of the local reflection Gaussians $\mathcal{G}_{\mathrm{local}}$ exhibits a concave structure extending inward from the physical surface, explicitly modeling the virtual reflection geometry. Without such decoupling, the geometry representation would be forced to collapse inward to match the virtual depth to reproduce specular appearance, leading to degraded surface reconstruction. By separating near-field specular reflection from geometry, our approach preserves surface correctness while enabling high-fidelity specular rendering.
Moreover, by avoiding explicit ray tracing, our approach maintains fast training speed and enables robust novel view synthesis, since mirror-induced virtual images are explicitly modeled as local Gaussians located behind the physical surface and shift naturally across the image with viewpoint changes.

\subsubsection{Physically-Aware Specular Adaptive Mixing Shader}
\label{sec:shader}
We predict specular radiance using a neural shader that combines two key ingredients: an adaptive mixing mechanism that balances global and local specular features, and physically-aware conditioning that enforces view-dependent reflectance behavior.

\paragraph{Adaptive mixing of global and local specular features.}
We provide the shader with two complementary specular features that characterize specular reflection at different spatial scales. The global specular features $\mathbf{F}_{\mathrm{global}}(\mathbf{p})$ capture far-field environment reflections, while the local specular features $\mathbf{F}_{\mathrm{local}}(\mathbf{p})$ account for near-field effects such as self-reflections and geometry-induced inter-reflections. Since their relative importance varies spatially and across viewpoints (e.g., open regions are dominated by environmental reflections, whereas concavities and self-occluding areas rely more on local reflections), the shader learns a view- and material-dependent mixing strategy evaluated per pixel. This adaptive mixing is learned implicitly through the joint utilization of $\mathbf{F}_{\mathrm{global}}(\mathbf{p})$ and $\mathbf{F}_{\mathrm{local}}(\mathbf{p})$ in the final specular radiance prediction.

\paragraph{Physically-aware conditioning.}
We take physically motivated geometric and material terms as explicit conditioning inputs, as they capture the key factors governing view-dependent reflection.
Specifically, we use the surface roughness $\rho(\mathbf{p})$ and the cosine between the surface normal $\mathbf{n}$ and the view direction $\mathbf{v}$,
\begin{equation}
\cos_{NV} = \max(\mathbf{n}\cdot \mathbf{v}, 0), 
\label{eq:cos_terms}
\end{equation}
which are concatenated with the specular features to guide the shading process.

\paragraph{Specular shading.}
The final specular radiance is predicted by a single MLP $f_{\Theta}$ that takes as input the global and local specular features together with the physically-aware conditioning terms:
\begin{equation}
\mathbf{C}_\mathrm{spec}(\mathbf{p}) = f_{\Theta}\big(
\mathbf{F}_{\mathrm{global}}(\mathbf{p}),\;
\mathbf{F}_{\mathrm{local}}(\mathbf{p}),\;
\rho(\mathbf{p}),\;
\cos_{NV}
\big).
\label{eq:mlp}
\end{equation}

Given calibrated multi-view images $\{\mathbf{I}_v\}$ and known camera parameters, the pipeline renders a diffuse (base) component from the geometry Gaussians $\mathcal{G}_{\mathrm{geo}}$ and a specular component predicted by the shader. The final linear color at pixel $\mathbf{p}$ is given by
\begin{equation}
\mathbf{C}(\mathbf{p}) =
\mathbf{C}_{\mathrm{diff}}(\mathbf{p}) + \mathbf{C}_{\mathrm{spec}}(\mathbf{p}).
\label{eq:final_color}
\end{equation}
For visualization, the linear RGB output is converted to sRGB by clamping to $[0,1]$ followed by the standard transfer function:
\begin{equation}
\mathbf{C}_{\mathrm{rgb}}(\mathbf{p})=
\operatorname{sRGB}\!\left(\operatorname{clamp}\!\left(\mathbf{C}(\mathbf{p}),0,1\right)\right),
\label{eq:spec_radiance}
\end{equation}
where $\operatorname{sRGB}(\cdot)$ denotes the canonical linear-to-sRGB conversion.

\subsection{Discussion}

\subsubsection{NeRFReN~\cite{guo2022nerfren}}

We discuss the differences between our method and NeRFReN. Although both introduce an extra scene representation to account for reflections, the two approaches differ substantially as follows:

\begin{itemize}
    \item Capability: While both can address virtual images, NeRFReN’s second scene is designed under the planar-mirror assumption. In contrast, our local reflection Gaussians $\mathcal{G}_{\mathrm{local}}$ model general near-field specular reflections (e.g., the interreflection between the bear’s face and the tabletop in the red boxed regions of Fig.~\ref{fig:glossyreal}, and the interreflection between the teapot spout and body in Fig.~\ref{fig:nvs}), where the virtual-image case is only one special instance.

    \item Method: While both introduce an extra scene, they represent different information, and NeRFReN models two components, whereas we model three. 
    NeRFReN decomposes color into transmission (standard volumetric rendering) and reflection; the reflection renders mirror-like reflections (i.e., planar reflectors) as a residual correction to the primary transmission. 
    In contrast, our first Gaussians $\mathcal{G}_{\mathrm{geo}}$ capture diffuse appearance and geometry, and query an environment map for far-field specular reflections; our second Gaussians $\mathcal{G}_{\mathrm{local}}$ represent general near-field specular reflections, not limited to planar reflectors. We model the final color as three components: diffuse, far-field specular, and near-field specular, leading to a fundamentally different color decomposition and modeling approach.

    \item Advantage (Geometry): NeRFReN doesn't support surface reconstruction and feeds both transmission and reflection branches with the same xyz-encoded features, coupling reflection appearance with the scene representation. If this coupled design is applied, view-dependent specular reflections can be incorrectly absorbed into the geometry representation, leading to appearance-driven geometry artifacts (ablations in Figs.~\ref{fig:ab_geo}--\ref{fig:ab_mae}). In contrast, we fully decouple geometry from specular reflection by using separately optimized geometry Gaussians $\mathcal{G}_{\mathrm{geo}}$ and local reflection Gaussians $\mathcal{G}_{\mathrm{local}}$, and modeling far-field specular reflections via environment map queries. This isolates view-dependent specular reflection from geometry Gaussians, enabling high-quality results for both surface reconstruction and novel view synthesis, rather than overfitting images through geometry distortions.

    \item Advantage (Label-free): NeRFReN composites the image as $I = I_t + \beta I_r$ using a learned reflection fraction map $\beta$, which often requires extra supervision (e.g., labeled reflection fraction maps). Instead of a $\beta$-blend, we mix far-field and near-field specular reflection via a physically-aware specular adaptive mixing shader (Sec.~\ref{sec:shader}). Moreover, our deferred rendering enforces global specular consistency through geometry-based environment queries, significantly reducing ambiguities and eliminating the need for manual labels.
\end{itemize}

\subsubsection{Ref-GS~\cite{zhang2025ref}}

\begin{figure*}[t]
\centering
\includegraphics[width=\textwidth]{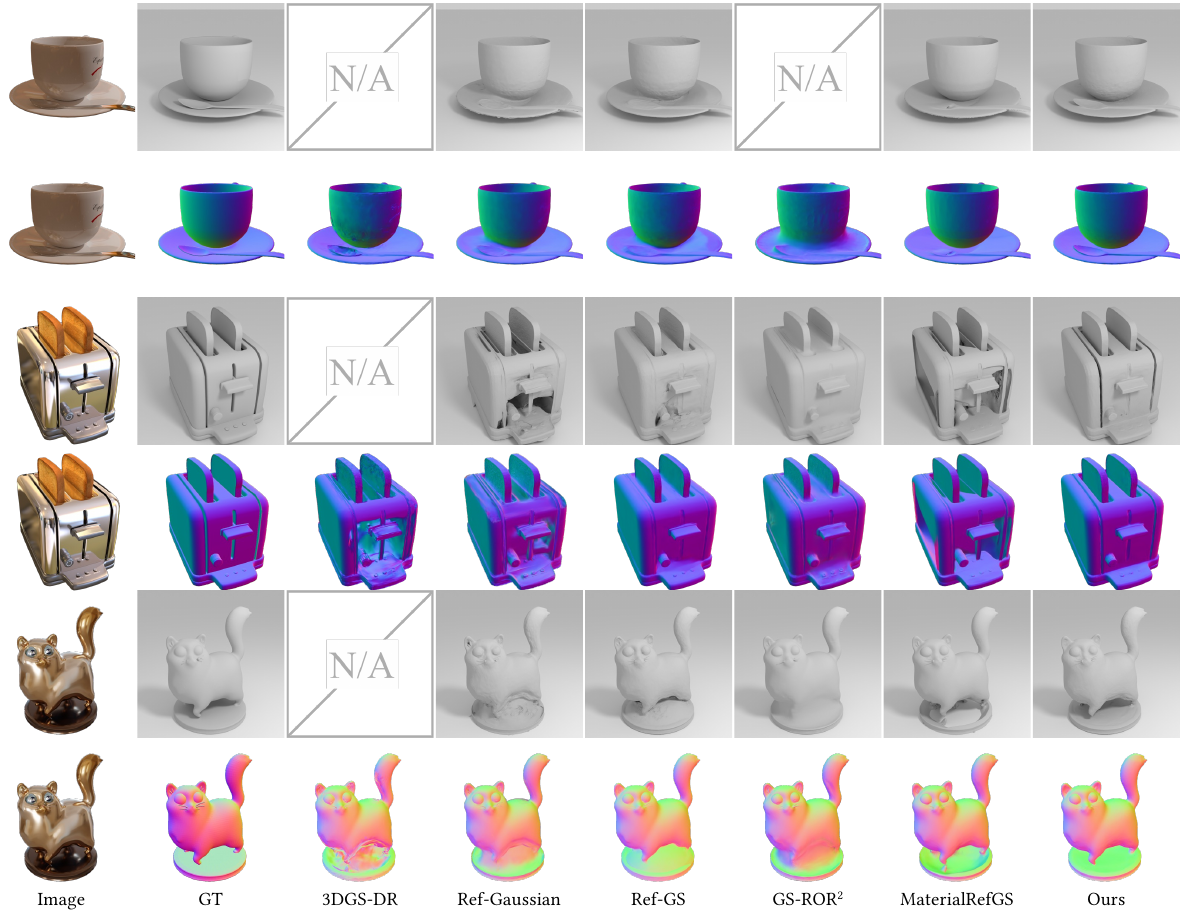}
\vspace{-0.7cm}
\caption{Qualitative surface reconstruction results (meshes and normals) on the ShinySynthetic (coffee, toaster) and GlossySynthetic (cat) datasets.}
\label{fig:geo}
\end{figure*}

\begin{figure*}[t]
\centering
\includegraphics[width=\textwidth]{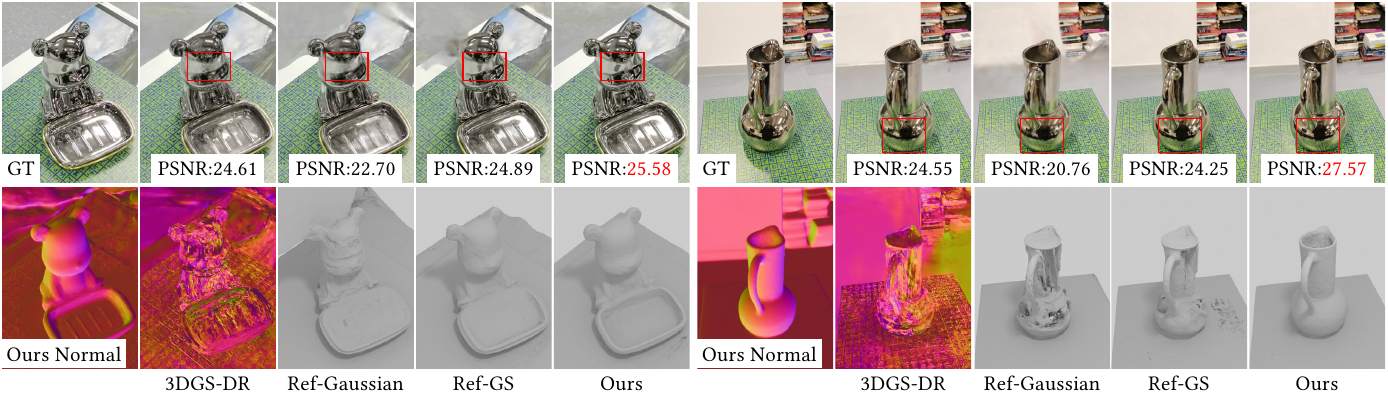}
\vspace{-0.7cm}
\caption{Qualitative rendering quality and surface reconstruction on the GlossyReal (bear, vase) dataset.}
\vspace{-0.3cm}
\label{fig:glossyreal}
\end{figure*}

\begin{figure*}[t]
\centering
\includegraphics[width=\textwidth]{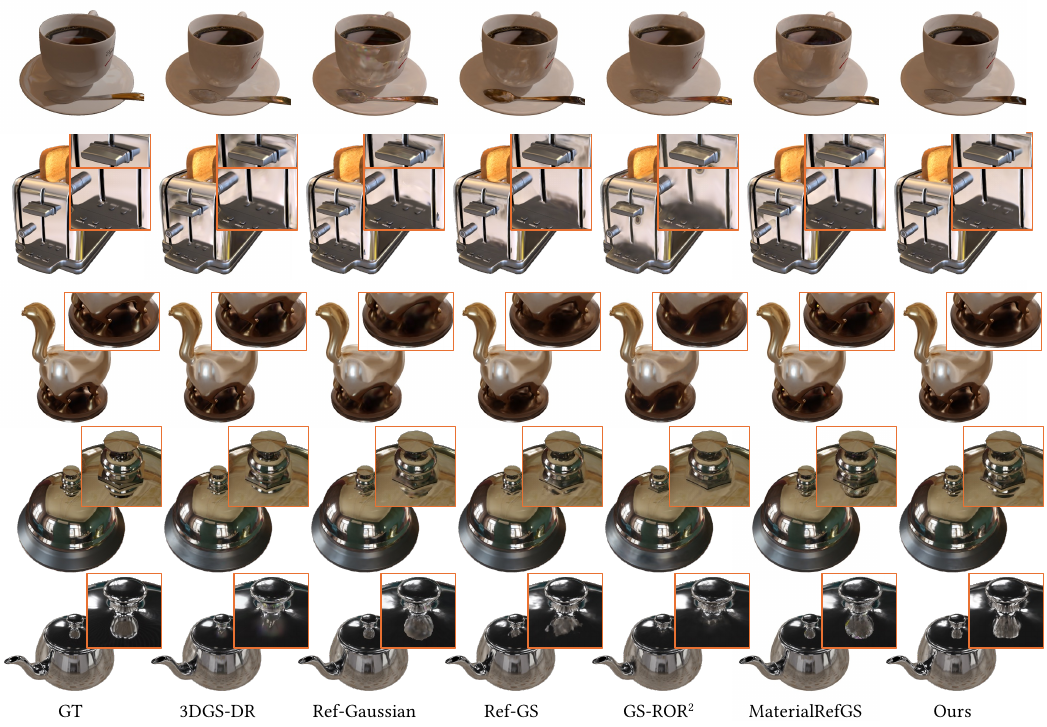}
\vspace{-0.7cm}
\caption{Qualitative NVS results on the ShinySynthetic (coffee, toaster) and GlossySynthetic (cat, tbell, teapot) datasets.}
\vspace{-0.2cm}
\label{fig:nvs}
\end{figure*}

Although the results of Ref-GS may appear to capture some near-field reflective effects (Fig.~\ref{fig:nvs}), Ref-GS does not explicitly model near-field specular reflections. Instead, near-field specular reflections and other view-dependent physical effects are absorbed into the spatial feature $\mathbf{K}$ and encoded using a single set of Gaussians, which entangles view-dependent appearance with the geometry-carrying Gaussians. Such entanglement can destabilize geometry optimization, causing view-dependent reflective effects to be incorrectly absorbed into geometry and resulting in reflection-induced geometry artifacts in regions with near-field specular reflections (Figs.~\ref{fig:geo}--\ref{fig:glossyreal}).

\begin{table}[t]
\caption{Quantitative surface reconstruction comparison on the ShinySynthetic dataset (normal MAE and training time). Best results are highlighted in \cellcolor{firstred}{red} and second-best in \cellcolor{secondyellow}{yellow}.}
\centering
\scriptsize
\setlength{\tabcolsep}{4pt}
\renewcommand{\arraystretch}{1.2}
\definecolor{firstred}{HTML}{F4C7C3}
\definecolor{secondyellow}{HTML}{FFF2CC}
\begin{tabular}{l|ccccccc|c}
\hline
 & ball & car & coffee & helmet &
 teapot & toaster & \textbf{mean} & time$\downarrow$\\
\hline
Ref-NeRF      & 1.55 & 14.93 & 12.24 & 29.48 & 9.23 & 42.87 & 18.38 & -- \\
ENVIDR        & 0.74 & 7.10 & 9.23 & 1.66  & 2.47 & 6.45 & 4.61 & -- \\
UniSDF        & \cellcolor{firstred} 0.45 & 6.88 & 8.00 & 1.72  & 2.80 & 6.45 & 8.71 & -- \\
PGSR& 66.93 & 4.62 & 2.91 & 6.01  & 1.01 & 15.31 & 16.13 & 27.5m \\
MILo& 61.30 & 4.47 & 2.36 & 4.87  & 0.62 & 16.49 & 15.02 & \cellcolor{secondyellow} 18.9m \\
GaussianShader& 7.03 & 14.05 & 14.93 & 9.33  & 7.17 & 13.08 & 10.93 & 1.10h \\
3DGS-DR      & 0.85 & 2.32 & 2.21 & 1.67 &  0.53 & 6.99 & 2.43 & 26.5m \\
Ref-Gaussian &  0.71 & 1.91 & 2.34 & 1.85  & \cellcolor{secondyellow} 0.48 & 5.70 & \cellcolor{secondyellow} 2.17 & 35.9m \\
Ref-GS        & 1.05 & 2.02 & 3.61 & 1.99 & 0.69 & \cellcolor{secondyellow} 3.92 & 2.21 & 23.5m \\
GS-ROR$^2$ & \cellcolor{secondyellow} 0.47 & 2.06 & 5.47 & 1.82  & 0.52 & 5.52 & 2.64 & 1.34h \\
MaterialRefGS        & 0.63 & \cellcolor{secondyellow} 1.86 & \cellcolor{firstred} 1.79 & \cellcolor{secondyellow} 1.46 & \cellcolor{firstred} 0.44 & 7.02 & 2.20 & 2.42h \\
Ours   & 0.61 & \cellcolor{firstred} 1.72 & \cellcolor{secondyellow} 1.86 & \cellcolor{firstred} 1.35 &  0.58 & \cellcolor{firstred} 2.43 & \cellcolor{firstred} 1.43 & \cellcolor{firstred} 12.6m \\ 
\hline
\end{tabular}
\label{tab:shiny_geo}
\vspace{-0.2cm}
\end{table}

\begin{table}[t]
\caption{Quantitative surface reconstruction comparison on the GlossySynthetic dataset (CD$\times 10^2$ and normal MAE). Best results are highlighted in \cellcolor{firstred}{red} and second-best in \cellcolor{secondyellow}{yellow}.}
\centering
\scriptsize
\setlength{\tabcolsep}{4pt}
\renewcommand{\arraystretch}{1.2}
\definecolor{firstred}{HTML}{F4C7C3}
\definecolor{secondyellow}{HTML}{FFF2CC}
\begin{tabular}{l|ccccccccc}
\hline
 & angel & bell & cat & horse &
 luyu & potion & tbell & teapot & \textbf{mean} \\
\hline
\multicolumn{10}{c}{CD$\downarrow$} \\
\hline
PGSR& 0.77 & 3.08 & 3.39 & 0.90  & 1.16 & 3.99 & 4.59 & 4.57 & 2.81 \\
MILo& 0.77 & 16.52 & 2.84 & 1.02  & 44.64 & 3.39 & 5.08 & 4.40 & 9.83 \\
GaussianShader& 0.85 & 1.10 & 2.56 & 0.73  & 1.07 & 4.74 & 5.74 & 3.40 & 2.53 \\
Ref-Gaussian & 0.45 & 0.70 & 1.68 & 0.64  & \cellcolor{secondyellow} 0.88 & 0.81 & 0.59 & 1.01 & 0.95  \\
Ref-GS        & \cellcolor{secondyellow} 0.41 & 0.74 & 1.73 & 0.47 & 0.89 & 1.05 & \cellcolor{secondyellow} 0.52 & 0.88 & 0.84\\
GS-ROR$^2$        & 0.52 & \cellcolor{firstred} 0.32 & \cellcolor{secondyellow} 1.66 & \cellcolor{secondyellow} 0.46  & 0.92 & 0.86 & 0.58 & \cellcolor{firstred} 0.66 & \cellcolor{secondyellow} 0.75 \\
MaterialRefGS & 0.53 & 0.70 & 1.97 & 0.47 & 0.96 & \cellcolor{secondyellow} 0.62 & \cellcolor{firstred} 0.55 & 1.02 & 0.85\\
Ours   & \cellcolor{firstred} 0.38 & \cellcolor{secondyellow} 0.65 & \cellcolor{firstred} 0.93 & \cellcolor{firstred} 0.42 & \cellcolor{firstred} 0.64 & \cellcolor{firstred} 0.66 & \cellcolor{secondyellow} 0.52 & \cellcolor{secondyellow} 0.79 & \cellcolor{firstred} 0.62  \\ 
\hline
\multicolumn{10}{c}{MAE$\downarrow$} \\
\hline
PGSR& 4.90 & 6.39 & 8.43 & 5.97  & 5.91 & 11.12 & 12.62 & 6.28 & 7.70 \\
MILo& 4.61 & 11.52 & 5.55 & 6.52  & 5.97 & 10.45 & 13.16 & 7.09 & 8.11 \\
GaussianShader& 2.90 & 1.60 & 4.33 & 3.27  & 4.56 & 9.52 & 5.60 & 3.01 & 4.35 \\
3DGS-DR       & 4.46 & 4.53 & 4.64 & 6.59 & 5.65 & 4.44 & 5.39 & 3.38 & 4.89\\
Ref-Gaussian & \cellcolor{firstred} 1.79 & 1.16 & 3.15  & 4.03 & 3.15 & 3.04 & 2.02 & 1.20 & 2.44  \\
Ref-GS        & \cellcolor{secondyellow} 1.99 & 0.92 & 2.93 & \cellcolor{secondyellow} 3.18 & \cellcolor{secondyellow} 2.82 & 3.64 & 1.87 & 1.18 & 2.32\\
GS-ROR$^2$        & 2.09 & \cellcolor{secondyellow} 0.86 & 3.37 & 2.85 & 3.15 & 4.04 & 2.64 & \cellcolor{secondyellow} 1.04 & 2.51 \\
MaterialRefGS & 1.81 & \cellcolor{secondyellow} 0.86 & \cellcolor{secondyellow} 2.42 & \cellcolor{firstred} 3.16 & 3.07 & \cellcolor{secondyellow} 2.68 & \cellcolor{firstred} 1.77 & 1.18 & \cellcolor{secondyellow} 2.12\\
Ours   & 2.05 & \cellcolor{firstred} 0.71 & \cellcolor{firstred} 1.37 & 3.44 & \cellcolor{firstred} 2.30 & \cellcolor{firstred} 2.48 & \cellcolor{secondyellow} 1.84 & \cellcolor{firstred} 0.86 & \cellcolor{firstred} 1.88 \\
\hline
\end{tabular}
\label{tab:glossy_geo}
\end{table}

\begin{table}[t]
\caption{Quantitative NVS comparison. Best results are highlighted in \cellcolor{firstred}{red} and second-best in \cellcolor{secondyellow}{yellow}.}
\centering
\scriptsize
\setlength{\tabcolsep}{2.5pt}
\renewcommand{\arraystretch}{1.2}
\definecolor{firstred}{HTML}{F4C7C3}
\definecolor{secondyellow}{HTML}{FFF2CC}
\begin{tabular}{l|ccc|ccc|ccc}
\hline
\multicolumn{1}{l|}{\textbf{Datasets}} & 
\multicolumn{3}{c|}{\textbf{ShinySynthetic}} &
\multicolumn{3}{c|}{\textbf{GlossySynthetic}} &
\multicolumn{3}{c}{\textbf{RefReal}} 
\\
\hline
 & PSNR$\uparrow$ & SSIM$\uparrow$ & LPIPS$\downarrow$ &
 PSNR$\uparrow$ & SSIM$\uparrow$ & LPIPS$\downarrow$ &
 PSNR$\uparrow$ & SSIM$\uparrow$ & LPIPS$\downarrow$ \\
\hline
Ref-NeRF      & 33.32 & 0.956 & 0.110 & 25.65 & 0.905 & 0.112 & 23.62 & 0.646 & 0.239 \\
ENVIDR        & 32.88 & 0.969 & 0.072 & 29.06 & 0.947 & 0.060 & 23.80 & 0.606 & 0.332 \\
PGSR& 26.82 & 0.921 & 0.119 & 25.59 & 0.913 & 0.091 & 24.43 & \cellcolor{secondyellow} 0.696 & \cellcolor{firstred} 0.220 \\
MILo& 27.28 & 0.921 & 0.124 & 25.69 & 0.914 & 0.096 &23.87 & 0.643 & 0.291 \\
GaussianShader& 30.42 & 0.951 & 0.082 & 27.11 & 0.922 & 0.082 & 23.46 & 0.647 & 0.257 \\
3DGS-DR      & 33.94 & 0.971 & 0.059 & 29.36 & 0.951 & 0.055 & 23.80 & 0.659 & 0.236 \\
EnvGS      & 33.26 & 0.968 & 0.063 & 28.17 & 0.938 & 0.067 & \cellcolor{secondyellow} 24.55 & 0.671 & 0.243 \\
Ref-Gaussian & \cellcolor{secondyellow}34.92 & \cellcolor{secondyellow}0.973 & \cellcolor{secondyellow}0.056 & 29.61 & 0.954 & 0.054 & 23.95 & 0.661 & 0.280 \\
Ref-GS        & 34.80 & \cellcolor{secondyellow}0.973 & \cellcolor{secondyellow}0.056 & 30.37 & \cellcolor{secondyellow} 0.960 & \cellcolor{secondyellow} 0.048 & 23.93 & 0.655 & 0.267 \\
GS-ROR$^2$       & 32.16 & 0.963 & 0.069 & 28.96 & 0.952 & 0.056 & -- & -- & -- \\
MaterialRefGS & 34.16 & 0.972 & 0.058 & \cellcolor{firstred} 30.74 & \cellcolor{firstred} 0.961 & \cellcolor{firstred} 0.047 & -- & -- & -- \\
Ours   & \cellcolor{firstred} 35.21 
& \cellcolor{firstred}0.975 
& \cellcolor{firstred}0.053 
& \cellcolor{secondyellow}30.63 
& 0.958 
& 0.052 
& \cellcolor{firstred}24.89 
& \cellcolor{firstred}0.697
& \cellcolor{secondyellow}0.224 \\ 
\hline
\end{tabular}
\vspace{-0.3cm}
\label{tab:nvs}
\end{table}

\section{Experiments}

\subsection{Implementation Details}

We jointly optimize the two sets of Gaussians with associated features, following the same optimization and densification strategy as 2D Gaussian splatting (2DGS)~\cite{huang20242d}. The Sph-Mip encoding uses a base level $\mathbf{M}$ with resolution $H_\mathbf{M}=512$, $W_\mathbf{M}=1024$, and feature dimension $d=4$, with $N=9$ mipmap levels. Global and local specular features are defined in $\mathbb{R}^{d}$ using the same feature dimension. The physically-aware specular adaptive mixing shader is implemented as a lightweight MLP $f_{\Theta}$ with 3 hidden layers of width 64. All experiments are conducted on a single NVIDIA RTX 4090 GPU with 24\,GB of memory.

\subsection{Datasets, Baselines and Metrics}
\subsubsection{Datasets.} 
We evaluate our method on two synthetic and two real-world datasets featuring glossy materials and complex specular transport.
For synthetic experiments, we use ShinySynthetic~\cite{verbin2022refnerf} and GlossySynthetic~\cite{liu2023nero}, which feature pronounced view-dependent reflections, high-frequency highlights, and strong near-field interactions, including self-reflections and inter-reflections. 
We additionally report results on the RefReal~\cite{verbin2022refnerf} and GlossyReal~\cite{liu2023nero} datasets to assess performance on real captures of reflective objects and scenes. 
For the RefReal dataset, we utilize $1/4$ resolution for the gardenspheres and toycar scenes, and $1/8$ resolution for the sedan scene. 
For the GlossyReal dataset, all experiments are conducted at $1/4$ resolution. 
Overall, these benchmarks present specular effects that are difficult to account for with a purely far-field environment model, making them well-suited for evaluating our global--local specular model.

\subsubsection{Baselines.}
We compare our method against state-of-the-art baselines from both neural volume rendering and 3D Gaussian splatting (3DGS). For neural volume rendering, we include Ref-NeRF~\cite{verbin2022refnerf}, ENVIDR~\cite{liang2023envidr}, and UniSDF~\cite{wang2024unisdf}. For 3DGS-based methods, we evaluate PGSR~\cite{chen2024pgsr}, MILo~\cite{guedon2025milo}, GaussianShader~\cite{Jiang_2024_CVPR}, 3DGS-DR~\cite{ye20243d}, EnvGS~\cite{Xie_2025_CVPR}, Ref-Gaussian~\cite{ICLR2025_abf3682c}, Ref-GS~\cite{zhang2025ref}, GS-ROR$^2$~\cite{zhu_2025_gsror}, and MaterialRefGS~\cite{zhangmaterialrefgs}. MaterialRefGS~\cite{zhangmaterialrefgs} is included as a baseline, reproducing results strictly from the publicly available code at the time of submission.

\subsubsection{Metrics.} 
We assess geometric accuracy using the mean angular error (MAE) of surface normals and the chamfer distance (CD) between reconstructed and reference surfaces, and evaluate novel view synthesis quality using PSNR, SSIM, and LPIPS.

For geometry evaluation, we report only normal MAE on ShinySynthetic~\cite{verbin2022refnerf} and both CD and normal MAE on GlossySynthetic~\cite{liu2023nero}. As discussed in Ref-NeuS~\cite{ge2023ref}, the ground-truth meshes of ShinySynthetic contain two surface layers, while only the outer layer is visible, making completeness-based metrics such as CD unreliable due to invisible inner surfaces. For GlossySynthetic, since ground-truth surface normals are not provided, normal MAE is computed using normals derived from the ground-truth depth.
Since the official implementation of GS-ROR$^2$~\cite{zhu_2025_gsror} does not include geometry evaluation code, we apply a unified evaluation pipeline across all methods. Normal MAE and CD follow our evaluation pipeline; for TSDF-based mesh extraction, we use a voxel size of $0.002$ when computing CD.

\begin{figure*}[h]
\centering
\includegraphics[width=0.93\linewidth]{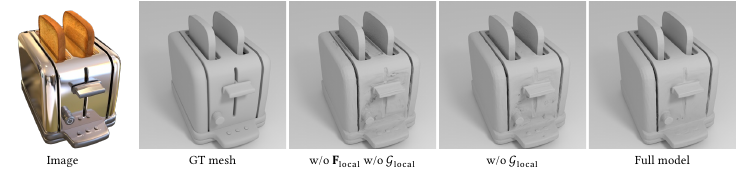}
\vspace{-0.4cm}
\caption{Ablation study on surface reconstruction.}
\label{fig:ab_geo}
\end{figure*}

\begin{figure}[t]
\centering
\includegraphics[width=\linewidth]{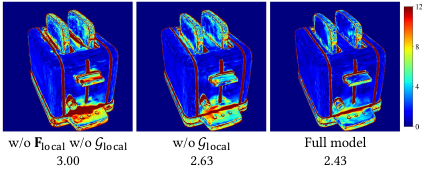}
\vspace{-0.7cm}
\caption{Ablation study on surface normal estimation. Angular error maps (red: high, blue: low) show that the full model significantly reduces geometric artifacts compared to ablated variants. Chamfer distances (CD) are reported at the bottom.}
\vspace{-0.3cm}
\label{fig:ab_mae}
\end{figure}

\begin{figure*}[h]
\centering
\includegraphics[width=0.95\linewidth]{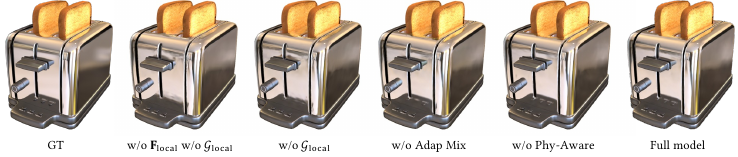}
\vspace{-0.4cm}
\caption{Ablation study on novel view synthesis.}
\vspace{-0.2cm}
\label{fig:ab_nvs}
\end{figure*}

\begin{figure}[t]
\centering
\includegraphics[width=\linewidth]{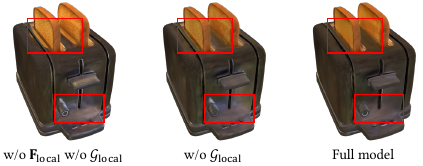}
\vspace{-0.7cm}
\caption{Ablation study on the diffuse leakage issue, visualizing the predicted diffuse color.}
\vspace{-0.2cm}
\label{fig:ab_diff}
\end{figure}

\subsection{Comparisons}

\begin{table}[t]
\caption{Rendering speed (FPS) on the ShinySynthetic dataset.}
\centering
\setlength{\tabcolsep}{8pt}
\renewcommand{\arraystretch}{1.2}
\begin{tabular}{l|cccc}
\hline

& EnvGS & MaterialRefGS & Ref-GS & Ours \\
\hline
FPS$\uparrow$ & 26.22 & 30.07 & 67.63 & \textbf{76.34} \\

\hline
\end{tabular}
\vspace{-0.2cm}
\label{tab:fps}
\end{table}

\subsubsection{Surface Reconstruction}

We evaluate surface reconstruction quality on all datasets. Tab.~\ref{tab:shiny_geo} reports the normal MAE and training time on the ShinySynthetic~\cite{verbin2022refnerf} dataset. Our method achieves the lowest reconstruction error while also exhibiting the shortest training time. Tab.~\ref{tab:glossy_geo} presents the CD and normal MAE on the GlossySynthetic~\cite{liu2023nero} dataset, where our method also outperforms competing approaches.
Fig.~\ref{fig:geo} shows qualitative surface reconstruction results on both ShinySynthetic~\cite{verbin2022refnerf} and GlossySynthetic~\cite{liu2023nero} datasets. Our method reconstructs sharp and well-defined boundaries on thin structures, such as the edges of the spoon, and avoids the common failure of merging the coffee cup with its base caused by occlusion-induced shadows and near-field reflections. In the toaster scene, which contains large front-facing surfaces dominated by near-field inter-reflections, our method produces smooth and coherent geometry without the surface artifacts observed in competing methods. Similarly, in the cat scene, our approach prevents incorrect depth and normal estimation induced by inter-object light transport, successfully separating the object from the base while faithfully reconstructing fine structures such as the cat’s thin whiskers.
Fig.~\ref{fig:teaser} and Fig.~\ref{fig:glossyreal} present qualitative surface reconstruction results on the RefReal~\cite{verbin2022refnerf} and GlossyReal~\cite{liu2023nero} datasets, respectively. In these real-world scenes with complex illumination and strong specular effects, our method reconstructs clean and consistent surfaces, preserving fine geometric details while avoiding spurious artifacts and surface distortions observed in competing methods.

A key challenge in the GlossyReal scenes, particularly the vase scene, is the presence of large specular reflections of nearby moving objects (e.g., the camera operator) observed on object surfaces. As these objects move across viewpoints, the effective illumination changes with view. Existing methods often misinterpret such view-dependent illumination as geometric cues, leading to spurious surface deformations, such as indentations aligned with reflected silhouettes. In contrast, our method remains robust under these conditions and reconstructs reliable surfaces. This robustness can be attributed to our local reflection Gaussians $\mathcal{G}_{\mathrm{local}}$ and their associated local specular features, which, beyond modeling near-field specular reflections, also adapt to complex, view-dependent variations in the environment illumination, preventing such effects from being incorrectly absorbed into the geometry.

\subsubsection{Novel View Synthesis}
We evaluate novel view synthesis performance on all datasets. Quantitative results are reported in Tab.~\ref{tab:nvs}, where our method consistently outperforms competing approaches. Fig.~\ref{fig:nvs} shows qualitative results. In scenes containing strong near-field specular reflections, competing methods fail to reproduce view-consistent appearance, leading to blurred or distorted regions, while our method preserves coherent appearance under novel viewpoints. Moreover, our method effectively handles curved near-field specular reflections, such as those between the cup and spoon in the coffee scene, and between the spout and body in the teapot scene, rather than being limited to planar reflective surfaces.
We do not report novel view synthesis results of GS-ROR$^2$~\cite{zhu_2025_gsror} and MaterialRefGS~\cite{zhangmaterialrefgs} on the RefReal~\cite{verbin2022refnerf} dataset, as GS-ROR$^2$ does not provide runnable code for this dataset, and the released implementation of MaterialRefGS is incomplete at the time of submission.

\subsubsection{Efficiency}
We report the average training time on the ShinySynthetic dataset in Tab.~\ref{tab:shiny_geo}. In addition, Tab.~\ref{tab:fps} compares the rendering speed (FPS) of our method with two ray-tracing-based methods, EnvGS and MaterialRefGS, as well as Ref-GS. The results show that our method is 2--3$\times$ faster than the ray-tracing-based approaches and also faster than Ref-GS, while achieving better surface reconstruction and novel view synthesis quality (Tab.~\ref{tab:nvs} and Fig.~\ref{fig:nvs}).

\begin{table}[t]
\caption{Ablation study on the ShinySynthetic (PSNR, SSIM, LPIPS, and normal MAE) and GlossySynthetic (CD$\times 10^2$ and normal MAE) datasets.}
\centering
\scriptsize
\setlength{\tabcolsep}{5pt}
\renewcommand{\arraystretch}{1.2}
\begin{tabular}{l|cccc|cccc|cc}
\hline
& PSNR$\uparrow$ & SSIM$\uparrow$ & LPIPS$\downarrow$ & MAE$\downarrow$ &  CD$\downarrow$ & MAE$\downarrow$\\
\hline
w/o $\mathbf{F}_{\mathrm{local}}$ w/o $\mathcal{G}_{\mathrm{local}}$ & 34.36 & 0.971 & 0.057 & 1.53 & 0.82 & 1.99\\
w/o $\mathcal{G}_{\mathrm{local}}$ & 34.70 & 0.973 & 0.057 & 1.46 & 0.73 & 1.93\\
w/o Adap Mix & 34.98 & 0.974 & 0.054 & 1.46 & 0.64 & 1.92\\
w/o Phy-Aware & 35.02 & \textbf{0.975} & \textbf{0.053} & 1.45 & \textbf{0.62} & 1.90 \\
Full model & \textbf{35.21} & \textbf{0.975} & \textbf{0.053} & \textbf{1.43} & \textbf{0.62} & \textbf{1.88}\\ 

\hline
\end{tabular}
\vspace{-0.2cm}
\label{tab:ablation}
\end{table}

\subsection{Ablation Studies}

We conduct ablation studies on the ShinySynthetic~\cite{verbin2022refnerf} and GlossySynthetic~\cite{liu2023nero} datasets to validate the effectiveness of individual components. Quantitative ablation results are reported in Tab.~\ref{tab:ablation}. Qualitative results for surface reconstruction and novel view synthesis are shown in Fig.~\ref{fig:ab_geo} and Fig.~\ref{fig:ab_nvs}, respectively. Fig.~\ref{fig:ab_mae} shows the angular error maps for surface normal estimation. Fig.~13 shows the diffuse color decomposition results. Additional decomposition results for more scenes are shown in Fig.~3 of the supplementary material.

\subsubsection{Local Features.}
We ablate the local specular features by removing $\mathbf{F}_{\mathrm{local}}(\mathbf{p})$ and modeling specular appearance using only the global specular features $\mathbf{F}_{\mathrm{global}}(\mathbf{p})$. Accordingly, the local reflection Gaussians $\mathcal{G}_{\mathrm{local}}$ are removed in this setting. This ablation evaluates the contribution of explicit near-field specular reflection beyond far-field specular reflection. As shown in Tab.~\ref{tab:ablation} and Figs.~\ref{fig:ab_geo}--\ref{fig:ab_diff}, when comparing w/o $\mathbf{F}_{\mathrm{local}}$ w/o $\mathcal{G}_{\mathrm{local}}$ and w/o $\mathcal{G}_{\mathrm{local}}$, removing the local specular features causes near-field specular reflections to be absorbed into the diffuse component, resulting in degraded geometry reconstruction and less accurate modeling of near-field specular effects.

\subsubsection{Dual Gaussians.}
We ablate the dual Gaussian decomposition by removing the local reflection Gaussians $\mathcal{G}_{\mathrm{local}}$ and instead deriving the local specular features $\mathbf{F}_{\mathrm{local}}(\mathbf{p})$ directly from the geometry Gaussians $\mathcal{G}_{\mathrm{geo}}$. This setting uses a single set of Gaussians to represent both geometry and near-field specular reflections, without explicit decoupling. As shown in Tab.~\ref{tab:ablation} and Figs.~\ref{fig:ab_geo}--\ref{fig:ab_diff}, this variant exhibits increased near-field specular leakage, degraded geometry reconstruction, and reduced fidelity in modeling near-field specular reflections compared to the variant with two separate sets of Gaussians, which we attribute to optimization conflicts between geometric representation and near-field specular reflections when modeled within a single set.

\subsubsection{Adaptive Mixing.}
We ablate the adaptive mixing mechanism in the specular shader by removing the learned global--local balancing and instead combining the global and local specular features by direct summation:
$\mathbf{F}_{\mathrm{sum}}(\mathbf{p}) =
\mathbf{F}_{\mathrm{global}}(\mathbf{p}) +
\mathbf{F}_{\mathrm{local}}(\mathbf{p})$.
Specular radiance is then predicted from $\mathbf{F}_{\mathrm{sum}}(\mathbf{p})$ together with the physical conditioning terms,
$\mathbf{C}_\mathrm{spec}(\mathbf{p}) =
f_{\Theta}\big(\mathbf{F}_{\mathrm{sum}}(\mathbf{p}),\,\rho(\mathbf{p}),\,\cos_{NV}\big)$.
As shown in Fig.~\ref{fig:ab_nvs}, direct feature summation yields unstable view-dependent specular responses. This is because, for a given view, specular appearance is typically dominated by either far-field or near-field specular reflections, depending on visibility, rather than an equal contribution of both. Moreover, the relative importance of these components varies across viewpoints. Without adaptive mixing, direct summation fails to model this view-dependent effect, leading to reduced generalization to novel views.

\subsubsection{Physically-Aware Conditioning.}
We ablate the physically-aware conditioning of the specular shader by removing all explicit physical terms from its inputs. Specifically, the roughness $\rho(\mathbf{p})$ and angular cosine term $\cos_{NV}$ are discarded, and specular radiance is predicted solely from the global and local specular features:
$\mathbf{C}_\mathrm{spec}(\mathbf{p}) =
f_{\Theta}\big(\mathbf{F}_{\mathrm{global}}(\mathbf{p}), \mathbf{F}_{\mathrm{local}}(\mathbf{p})\big).$
As shown in Tab.~\ref{tab:ablation} and Fig.~\ref{fig:ab_nvs}, removing physical conditioning results in unstable and less plausible view-dependent specular behavior, indicating that explicit physical terms help regularize optimization and improve generalization to novel views. The improvement is relatively modest, likely because the expressive shader MLP and high-dimensional specular features can implicitly encode some view-dependent and roughness-related information through multi-view Gaussian splatting.

\section{Conclusion}

We have presented \textbf{Ref-DGS}, a reflective dual Gaussian splatting framework that enables efficient and high-quality modeling of reflective appearance without explicit ray tracing. The key insight of Ref-DGS is to decouple near-field specular reflection from surface geometry via a dual Gaussian scene representation, assigning near-field reflective interactions to a dedicated set of local reflection Gaussians while preserving stable and accurate surface reconstruction. By representing specular appearance using complementary global and local specular features and predicting specular radiance with a lightweight, physically-aware specular adaptive mixing shader, Ref-DGS captures both far-field and near-field specular reflections within a rasterization-based pipeline.

Ref-DGS achieves state-of-the-art performance on reflective scenes in both surface reconstruction and novel view synthesis, while training substantially faster than ray-based Gaussian methods. Beyond improved visual/geometric quality and efficiency, our model demonstrates that explicitly separating near-field specular reflection from surface geometry is critical for stable and accurate modeling of complex view-dependent appearance. We believe this dual-representation perspective provides a useful insight for future extensions of Gaussian splatting toward more expressive modeling of view-dependent appearance.

\begin{acks}
This work was supported by the National Natural Science Foundation of China (62202076 and 52238003) and the Key Project of the National Natural Science Foundation of China (12494550 and 12494553).
\end{acks}

\bibliographystyle{ACM-Reference-Format}
\bibliography{sample-bibliography}

\end{document}